\documentclass[letterpaper]{article} 
\usepackage{aaai2026}  
\usepackage{times}  
\usepackage{helvet}  
\usepackage{courier}  
\usepackage[hyphens]{url}  
\usepackage{graphicx} 
\usepackage{natbib}  
\usepackage{caption} 
\usepackage{algorithm}
\usepackage{algorithmic}
\usepackage{amsmath}
\usepackage{amsthm,amsmath,amssymb}
\usepackage{mathrsfs}
\usepackage{newfloat}
\usepackage{listings}
\DeclareCaptionStyle{ruled}{labelfont=normalfont,labelsep=colon,strut=off} 
\floatstyle{ruled}
\newfloat{listing}{tb}{lst}{}
\floatname{listing}{Listing}
\title{Breaking the Passive Learning Trap: An Active Perception Strategy for Human Motion Prediction}
\author{
    Juncheng Hu\textsuperscript{\rm 1}, Zijian Zhang\textsuperscript{\rm 1}, Zeyu Wang\textsuperscript{\rm 2}, Guoyu Wang\textsuperscript{\rm 1}, Yingji Li\textsuperscript{\rm 1}, Kedi Lyu\textsuperscript{\rm 1}\thanks{Corresponding Author: Kedi Lyu}\\  
}
\affiliations{
    \textsuperscript{\rm 1}College of Computer Science and Technology, Jilin University, Changchun, China\\
    \textsuperscript{\rm 2}College of Computer Science and Engineering, Dalian Minzu University, Liaoning, China\\

    jchu@jlu.edu.cn, zhangzijian@jlu.edu.cn, 20231578@dlnu.edu.cn, wgy21@mails.jlu.edu.cn, yingjili@jlu.edu.cn, kdlyu@jlu.edu.cn
}

\usepackage{bibentry}

\begin{document}

\maketitle

\begin{abstract}
Forecasting 3D human motion is an important embodiment of fine-grained understanding and cognition of human behavior by artificial agents.
Current approaches excessively rely on implicit network modeling of spatiotemporal relationships and motion characteristics, falling into the \textbf{\textit{passive learning trap}} that results in redundant and monotonous 3D coordinate information acquisition while lacking actively guided explicit learning mechanisms.
\quad To overcome these issues, we propose an \textbf{A}ctive \textbf{P}erceptual  \textbf{S}trategy (\textbf{APS}) for human motion prediction, leveraging quotient space representations to explicitly encode motion properties while introducing auxiliary learning objectives to strengthen spatio-temporal modeling.
\quad Specifically, we first design a \textit{data perception module} that projects poses into the quotient space, decoupling motion geometry from coordinate redundancy. By jointly encoding tangent vectors and Grassmann projections, this module simultaneously achieves geometric dimension reduction, semantic decoupling, and dynamic constraint enforcement for effective motion pose characterization.
\quad Furthermore, we introduce a \textit{network perception module} that actively learns spatio-temporal dependencies through restorative learning. This module deliberately masks specific joints or injects noise to construct auxiliary supervision signals. A dedicated auxiliary learning network is designed to actively adapt and learn from perturbed information.
Notably, APS is model agnostic and can be integrated with different prediction models to enhance \textit{active perceptual}.
The experimental results demonstrate that our method achieves the new state-of-the-art, outperforming existing methods by large margins: 16.3\% on H3.6M, 13.9\% on CMU Mocap, and 10.1\% on 3DPW.
\end{abstract}


\section{Introduction}

\begin{figure}[t]
  \includegraphics[width=.45\textwidth]{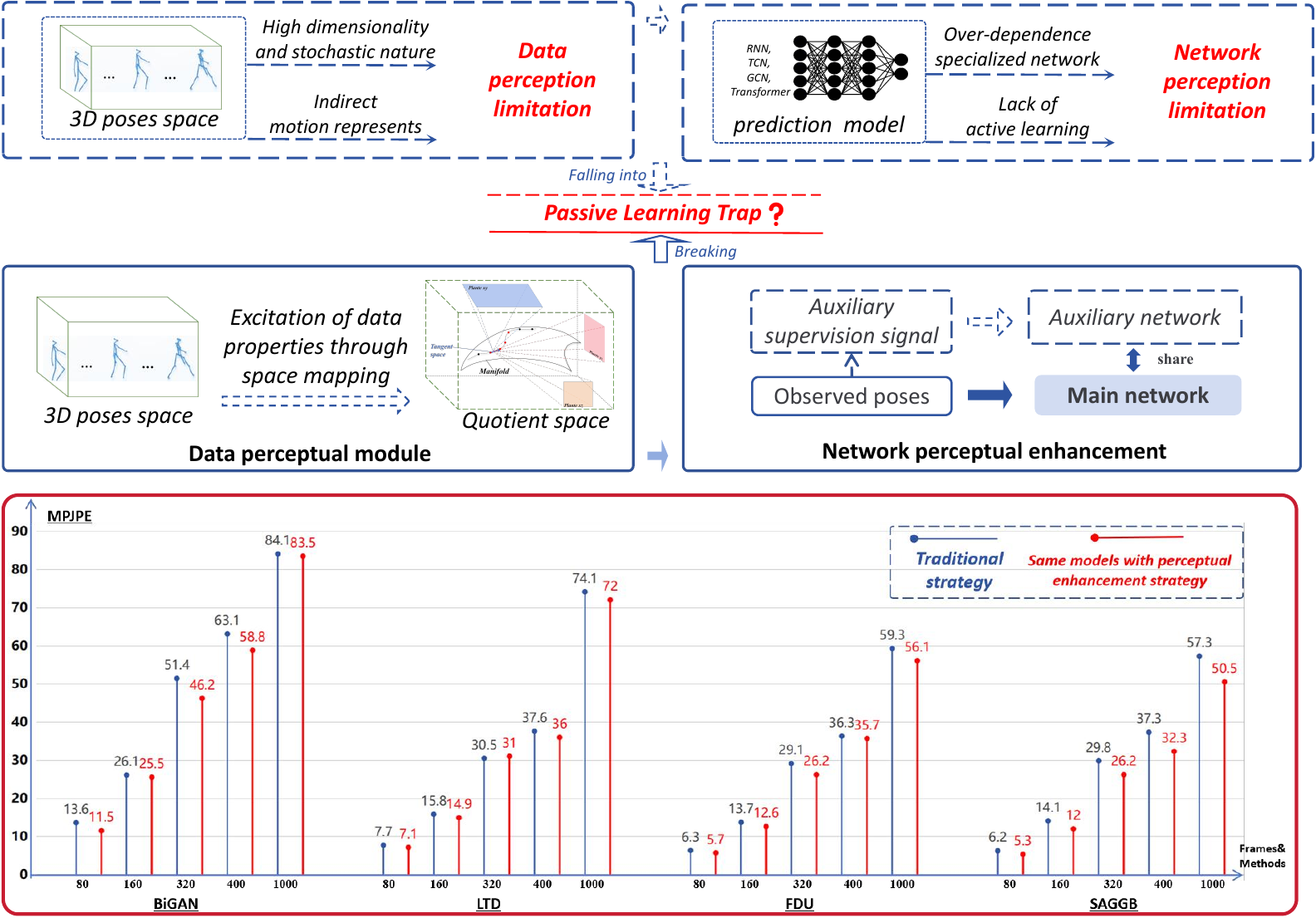}
  \centering
  \caption{
  The upper part highlights the motivation behind APS, while the lower part demonstrates its effectiveness in mitigating premature performance bottlenecks. 
  }
   \label{fig:teaser}
\end{figure}

Modeling 3D human motion from high-dimensional and highly stochastic historical observations to achieve accurate future human motion prediction (HMP) is extremely challenging.
Given its significant implications, HMP has received diverse applications, including embodied intelligence, human-computer interaction, and autonomous driving \cite{Hci,autov,cvpr/ZhouW23,iccv/YaoLSCLOL23}. 

In this paper, we focus on enhancing motion prediction through hierarchical perception (\textit{i.e.} \textit{spatio-temporal relationships} and \textit{motion properties}) and capture long-range spatio-temporal dependencies to produce accurate predictions. 
Recent methods propose specialized network structures to model human motion. Some approaches utilize RNNs \cite{liu2022} and TCNs \cite{sensor2023, Nips} to model temporal dependencies. Some methods \cite{KBS2024, TangZDGY23, cui2024, DBLP:conf/mm/ZhangZMWL23} propose GCN networks with learnable weights. 
DMGNN \cite{DMGNN} and MST-GNN \cite{PGS} further construct multi-scale body graphs to model local-global spatial features. 
PGBIG \cite{eccv/LiCZXTZ22} also designs temporal graph convolution extract spatio-temporal features. 
SPGSN \cite{HRI} proposes graph scattering networks to further model the temporal dependence from multiple graph spectral bands. 

Scrutinizing the released implementations of existing methods, one observes that current methods often suffer from limitations that lead to \textit{falling into the passive learning trap}, thus affecting prediction accuracy, as illustrated in Figure~\ref{fig:teaser}. 
\quad We believe that this issue is mainly due to the following two reasons:
\textit{\textbf{Firstly}}, \textit{data perception limitation}. 
Existing methods \cite{Res-gru,AMH} perform dynamic modeling in 3D pose space, inheriting the intrinsic complexity of human motion ($e.g.,$ high dimensionality and stochastic nature). In such high-dimensional regimes, effective motion features often become entangled, while many implicit kinematic properties resist effective encoding within rigid 3D representations. Without active guidance, the network will be forced to learn redundant coordinate system information.
\quad \textit{\textbf{Secondly}}, \textit{network perception limitation}. 
Current approaches \cite{MGCN,tcsvtrans} predominantly rely on passive information aggregation through static architectures, lacking an explicit, actively guided learning mechanism. This inherent limitation restricts their ability to dynamically adapt to complex spatiotemporal dependencies, ultimately impairing their effectiveness in modeling intricate motion patterns.

\begin{figure*}[h]
	\centering
	\includegraphics[width=\textwidth]{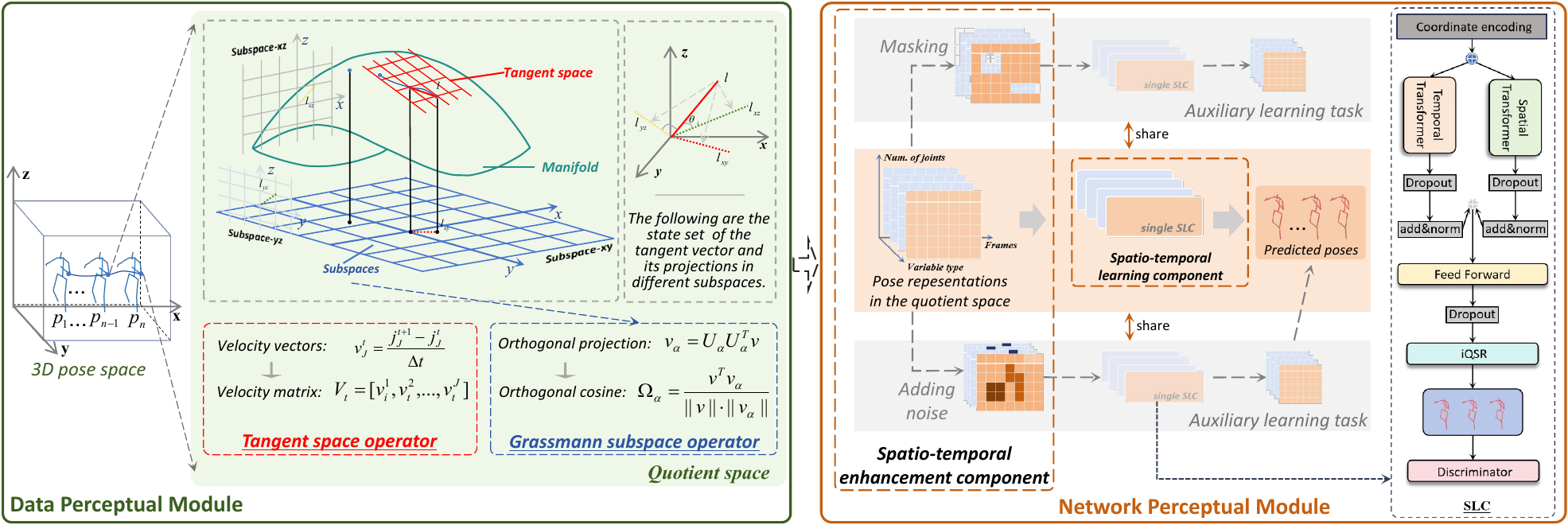}
	\caption{The architecture of the proposed method. }
	\label{fig:architecture}
\end{figure*}

To tackle these challenges, an active perceptual strategy (APS) for HMP is proposed to facilitate active perception of spatio-temporal dependencies and motion information to mitigate the undesired impact of \textbf{\textit{passive learning trap}}, as illustrated in Figure~\ref{fig:architecture}. APS consists of two major modules, namely a \textit{data perception module} (DPM) and a \textit{network perception module} (NPM). 

The \textbf{DPM} dresses high-dimensional human pose perception challenges via a geometric framework that projects pose sequences onto the quotient space, which includes two parts.
The \textit{tangent space} capturing local motion dynamics via tangent vectors $v$ representing instantaneous pose variations, and the \textit{Grassmann manifold} $Gr(k,n)$ embeddings that model low-dimensional subspace states ($e.g.,$ joint movement constraints). 
\quad Specifically, the motion of each joint can be regarded as a trajectory on the manifold $\mathcal{M}$, and the displacements between frames are the tangent vectors $v$ of this estimate. 
\quad \textit{First}, we project poses onto the tangent space , and utilize the tangent space operator to calculate the trajectory of joint motion, representing it as the tangent vector $v$.
\quad \textit{Then}, each vector is projected onto the Grassmann manifold, \textit{i.e.,} decomposed into three different fixed subspaces $Gr(2,3)$. The angle between the tangent vector and the coordinate plane reflects the relative orientation of $v$ with respect to these subspaces. The angle characterizes the orientation of the tangent vector on the Grassmann manifold, describing how $v$ is distributed in the different subspaces. 
\quad \textit{Finally}, these subspace-projected motion data are processed by NPM to generate future motion data. 

The \textbf{NPM} aims to address network perception limitation by constructing auxiliary supervision signals to force the network to actively repair the spatiotemporal relationships that are missing or disturbed by noise. 
NPM comprises two components.
A \textit{spatio-temporal enhancement component} (SEC) induces active perception by selectively corrupting observed pose to formulate auxiliary reconstruction tasks. Through reconstructing these masked data into their original configurations, it establishes auxiliary learning objectives that compel downstream networks to develop intrinsic motion pattern comprehension. These intentionally corrupted data are then fed into a \textit{spatial-temporal learning component} (SLC). 
To adapt to this dynamically changing human motion data format, we design SLC as a data learning component. SLC employs spatio-temporal graph attention mechanisms to dynamically integrate local relationships with global motion semantics. By iteratively refining the reconstruction of corrupted poses through adaptive feature aggregation across temporal and spatial dimensions, it achieves active perception of latent motion dependencies without predefined structural priors. The attention units automatically emphasize salient spatio-temporal  correlations while suppressing noise propagation, enabling the model to capture comprehensive motion dynamics through self-taught feature interactions.

\textbf{Contributions.} \quad To summarize, our key contributions are as follows:
\quad i)  We propose a novel active perception strategy that is model-agnostic for human motion prediction tasks.
\quad ii) We present a data perception module and a network perception module to separately enhance pose representation and dynamic context modeling, jointly enhancing the framewok’s active perception capability, and elevating prediction accuracy.
\quad iii) Our method achieves SOTA results on three benchmark datasets, 16.3\% on H3.6M, 13.9\% on CMU Mocap, and 10.1\% on 3DPW.

\section{Our Approach}
\textbf{Problem formulation.} \quad
Given poses sequence $P^n=\langle p_{0}, p_{1}, \ldots, p_{n}\rangle$, the network captures implicit spatio-temporal dependencies and predicts the future pose sequence $P^N=\langle \hat{p}_{n+1}, \hat{p}_{n+2}, \ldots, \hat{p}_{n+N}\rangle$. $p_{n} \in R^{J * 3}$ represents the human pose at the $n_{th}$ frame, and $J$ denotes the total number of joints. $N$ refer to the lengths of future movements.  The objective is to learn a prediction model $ F_{pred}(\cdot)$ such that the predicted future motion $\hat{P}^N=F_{pred}(P^n)$ is closed to the ground truth $P^N$. 
\quad Inspired by the motion in the quotient space, we propose decomposing and embedding the human pose sequence into different subspace constraints, whereby the original motion can be represented as a dynamic projection on the fiber tuft $\eta$. Formally, let $T_p\mathcal{M}$ describe the instantaneous motion ($e.g.,$ joint velocity or displacement) of a human pose manifold $\mathcal{M}$ at a point $p$ (a certain pose). $Gr(k,d)$ denotes the space of all possible $k$-dimensional projections in a $d$-dimensional tangent space. Thus, the fiber bundle constructed by the human pose sequence can be represented as:
\begin{equation}
\label{eq:1}
\eta=(T_p\mathcal{M},Gr(k,d)).
\end{equation}
At this point, we can describe the motion by learning the tangent vectors and their projections on subspaces.

\textbf{Method Overview.} 
\quad In this paper, we propose an \textbf{A}ctive \textbf{P}erception \textbf{S}trategy (APS) consisting of two key modules: a \textit{data perceptual module} (DPM) and a \textit{network perceptual module} (NPM), as illustrated in Figure~\ref{fig:architecture}.
Specifically, the DPM is \textit{first} designed to mitigate data perception limitations by transforming motion sequences from 3D pose space into quotient space. The processed data is \textit{then} fed into the NPM, where active noise injection or joint masking is applied to the raw skeletal data, thereby stimulating the network's active perception capability. \textit{Finally}, the perturbed yet structured data is input into the learning network for predicting future motion trajectories in subspace.
The following sections elucidate our proposed DPM and NPM in detail.

\subsection{Data Perceptual Module \label{DPE}}
As mentioned above, human motion is inherently characterized by high-dimensionality (\textit{e.g.,} multi-joint coordination) and stochastic nature (\textit{e.g.,} variable movement patterns). 
Existing methods typically model motions directly in high-dimensional pose spaces, forcing the network to learn redundant coordinate system information, and passively falling into the coupled representation of geometry and semantics.
\quad Based on this, we propose the \textbf{DPM} to utilize the quotient space representation to actively enhance the implicit dynamic information while reducing the high-dimensional poses. 
Specifically, by mapping to the quotient space, pose sequences are decomposed and embedded into subspace-constrained representations, where primitive motions are dynamically projected onto a fiber bundle. Formally, this reduces motion modeling to tracking tangent space variations $v$ on poses manifold $\mathcal{M} $ at point $p$ relative to its Grassmann structure $Gr(\cdot)$.
For engineering implementation, we design the \textit{tangent space operator} (TSO) and \textit{Grassmann manifold operator} (GMO) to complete the mapping.

\textbf{Tangent space operator.} \quad
The objective of TSO is to describe the motion change of human poses.
Let the human pose have $J$ joints, the time series length be $n$, and the original data be $P=\{p_t\}_{t=1}^n$ and $p_t=[j_t^1,...,j_t^J]$. For each joint $j$, the neighboring frame velocity vectors are computed:
\begin{equation}
\label{eq:state_operator}
v_J^t=\frac{j_J^{t+1}-j_J^{t}}{\Delta t}.
\end{equation}
Then, considering the velocity field as a tangent vector on the manifold $\mathcal{M}$, the velocity matrix $V_t=[v_t^1,v_t^2,...,v_t^J]$ can be constructed, and the tangent vector is expressed as:
\begin{equation}
\textbf{v}_t=vec(V_t).
\end{equation}
Thus, in the tangent space $T_p\mathcal{M}$, the human pose $p$ is represented as a set of tangent vectors $vec(\cdot)$.

\textbf{Grassmann manifold operator.}\quad
Different from the Lie-group based methods, which calculate the rotation of poses through complex systems such as exponential mapping, logarithmic mapping and Lie algebra, we design GMO to describe the orientation change of tangent vector in the subspaces. The tangent vector $v$ from TSO is projected onto the Grassmann manifold $Gr(k,n)$ ($k$-dimensional projections in a $d$-dimensional tangent space), which is decomposed into three different fixed subspaces $Gr(2,3)$, each subspace is a 2D space $\mathcal{S}$, which is represented by an orthonormal basis matrix $\mathbf{U_\alpha} = [\mathbf{u}_1, \dots, \mathbf{u}_n], \alpha\in \{xy,yz,zx\}$. 
For a subspace $\mathcal{S}_\alpha = \text{span}(\mathbf{U}_\alpha)$, orthogonal projection of $\mathbf{v}$ onto $\mathcal{S}$:
\begin{equation}
\label{eq:state_operator1}
\mathbf{v}_{\alpha} = \mathbf{U_{\alpha}}\mathbf{U}^\top_{\alpha} \mathbf{v} = \sum_{i=1}^k (\mathbf{v}^\top \mathbf{u}^i) \mathbf{u}^i.
\end{equation}
The orthogonal cosine $ \mathbf{\Omega}_\alpha$ between \(\mathbf{v}\) and its projection \(\mathbf{v}_\alpha\) is computed as:
\begin{equation}
\label{eq:state_operator2}
 \mathbf{\Omega}_{\alpha} = \frac{\mathbf{v}^\top \mathbf{v}_\alpha}{\|\mathbf{v}\|\cdot\|\mathbf{v_{\alpha}}\|}.
\end{equation}
The information from GMO is expressed through an orthogonal decomposition of the fine-grained dynamics rather than skeletal joint coordinates. 
The quotient space representation $\mathcal{Q}$ is denoted as
$\mathcal{Q}^i = \{{p}_{1}, \mathbf{v}_t, \mathbf{\Omega}^i\}$
, where ${p}_{1}$ denotes the last frame of the observations. Since the variables in GMO represent the rotation of the human pose, we only need to obtain the modulus of the tangent vector in TSO.

\subsection{Network Perceptual Module \label{NPE}}
Currently, our quotient space representation mitigates the challenges of high-dimensional stochasticity through three techniques: geometric dimensionality reduction, semantic decoupling, and dynamic constraints, thereby reducing the learning burden while improving data perception. 
\quad However, existing models rely excessively on networks to implicitly capture spatio-temporal dependencies, but lack an actively guided explicit learning mechanism.
Therefore, we design a network perceptual module (\textbf{NPM}) to construct an auxiliary supervision signal through adversarial perturbation (mask/noise injection), compelling the network to actively repair spatiotemporal relationships.
\quad Specifically, we \textit{firstly} design a \textit{spatio-temporal enhancement component} (\textbf{SEC}) that induces active perception by selectively destroying the observed attitude coordinates to formulate \textbf{\textit{the auxiliary learning task}}. 
This design imposes two requirements on the modeling network: i) the ability to reconstruct masked coordinates by learning spatio-temporal relationships between corrupted and intact data, and ii) robustness to incomplete motion sequences induced by the masking process. 
For these, we introduce a \textit{spatio-temporal learning component} (\textbf{SLC}), implemented as a generative adversarial transformer, which dynamically infers and adapts to complex motion patterns while recovering the perturbed structure.

\textbf{Spatio-temporal enhancement component.} \quad
In the auxiliary learning tasks, each coordinate $x_t^i$ in the historical motion sequence is randomly masked with probability $p_m$. The objective is to reconstruct these masked coordinates from the observed unmasked values. In another way, Gaussian noise \( \varepsilon \sim \mathcal{N}(0, \sigma) \) is added to each coordinate \( x_t^i \) with probability \( p_n \), where \( \sigma \) controls the noise intensity. The goal is to recover clean motion from these corrupted inputs.  
Let \( P \), \( P_M \), and \( P_D \) denote the original motion sequence, masked sequence, and noisy sequence. While all tasks share the same backbone network, they employ distinct prediction heads to specialize in their respective objectives.  


\textbf{Spatio-temporal learning component.} \quad
To learn incomplete motion data, we process each coordinate as an independent feature while modeling spatio-temporal dependencies through a coordinate-level attention mechanism. We introduce a learnable masking token to explicitly mark masked coordinates, seamlessly integrating this masking awareness into the attention mechanism to handle arbitrary missing values. Given the inherent complexity of disentangling and fusing such spatio-temporal information, we leverage a WGAN for training, capitalizing on its demonstrated effectiveness for prediction tasks. SLC maps DPM data to an embedding space via a linear encoder.


\textit{Spatio-temporal attention.}
To capture the distinct patterns in joint spatial correlations and temporal dynamics, our SLC decouples these dependencies via dedicated attention mechanisms. The spatial attention module processes features within each timestamp: for features \( H^t \in \mathbb{R}^{J \times D} \) at time \( t \), we project them into queries \( Q^t \), keys \( K^t \), and values \( V^t \) via linear transformations \( \delta(\cdot) \), then compute a masked spatial attention matrix \( A_s \in \mathbb{R}^{J \times J} \). Following recent efficient attention designs, we adopt a low-rank approximation:  
\begin{equation}
\label{equ1}
{head}^{t}=\varphi\left(Q^{t}\right)\left(\varphi(K)^{T} V^{t}\cdot A_{s}\right),
\end{equation}
where \( \varphi(\cdot) \) denotes a dimensionality-reducing nonlinear. The outputs of \( H \) are concatenated and processed by a feedforward network \( \phi(\|_{i=1}^H \text{head}_i^t) \).  
Conversely, the temporal attention module operates on joint-specific sequences: for features \( H_j \in \mathbb{R}^{T \times D} \) of joint \( j \), we derive \( Q_j \), \( K_j \), \( V_j \) analogously and compute weights \( A_t \in \mathbb{R}^{T \times T} \):  
\begin{equation}
\label{equ2}
{head}_{j}=\varphi\left(Q_{j}\right)\left(\varphi\left(K_{j}\right)^{T} V_{j}\cdot A_{t}\right).
\end{equation}

\textit{Motion discriminator.}
We employs a WGAN-GP architecture to improve training stability. The adversarial system consists of: a generator based on a spatio-temporal deformable Transformer, and two specialized discriminators. The fidelity discriminator ensures pose realism, while the continuity discriminator preserves temporal coherence.

\subsection{Inference Objective \label{fcpm}}
To accommodate diverse task objectives (prediction, masking, and denoising), our framework applies targeted supervision to distinct sequence segments. Let \( M \) denote masked joints, and \( \hat{p} \), \( \hat{p}_M \), \( \hat{p}_D \) represent outputs for prediction, masking, and denoising tasks respectively. The composite loss integrates task-specific terms: 
\begin{equation}
    \begin{split}
L = \underbrace{\frac{1}{T_f J} \sum_{t,j} \|\hat{p}_j^t - p_j^t\|^2}_{L_{pred}} + \alpha_1 \underbrace{\frac{1}{|M|} \sum_M \|\hat{p}_{M,j}^t - p_j^t\|^2}_{L_{mask}} \\
+ \alpha_2 \underbrace{\frac{1}{T_p J} \sum_{t,j} \|\hat{p}_{D,j}^t - p_j^t\|^2}_{L_{denoise}}.
    \end{split}
\end{equation}
where \( \alpha_1, \alpha_2 \) balance task contributions.  
For adversarial training, we adopt WGAN-GP to minimize the Wasserstein distance between generated (\( P_g \)) and real (\( P_r \)) sequence distributions:
\begin{equation}
\label{equ4}
    \begin{split}
        L_{adv}=E_{X^{\prime} \sim P_{g}}\left[D\left(x^{\prime}\right)\right]-E_{X \sim P_{r}}[D(x)] \\
        +\lambda E_{x^{\prime} \sim P_{{x}^\prime}}\left[\left(\left\|\nabla_{{x}^\prime} D\left(x^{\prime}\right)\right\|_{2}-1\right)^{2}\right],
    \end{split}
\end{equation}
where \( \hat{X} \) interpolates real and generated data, and \( \lambda \) controls gradient penalty strength. This penalty enforces Lipschitz continuity, stabilizing GAN training. The total loss combines all terms: 
\begin{equation}
\label{equ5}
    L_{g}=\beta_{1} L+\beta_{2} L_{adv}.
\end{equation}

\begin{table*}[t]
\centering
\resizebox{\textwidth}{!}{
\renewcommand\tabcolsep{6.0pt}
\begin{tabular}{c|cccccc|cccccc|cccccc} 
\hline
Time (ms) & 80           & 160           & 320           & 400           & 560           & 1000           & 80            & 160           & 320           & 400           & 560           & 1000           & 80            & 160           & 320           & 400           & 560           & 1000            \\ 
\hline
         & \multicolumn{6}{c|}{Phoning}                                                                  & \multicolumn{6}{c|}{Eating}                                                                    & \multicolumn{6}{c}{Purchases}                                                                   \\

HMR      & 12.5         & 21.3          & 39.3          & 58.6          & 71.3          & 112.8          & 9.2           & 13.9          & 34.6          & 47.1          & 61.3          & 84.8           & 15.3          & 30.6          & 64.7          & 73.9          & 97.5          & 122.7           \\
GA-MIN   & 8.3          & 17.8          & 37.9          & 44.8          & 63.0          & 101.5          & \textbf{5.8 }          & 12.5          & 25.3          & 33.8          & 47.3          & 65.2           & 12.4          & 28.5          & 60.0          & 72.9          & 89.9          & 135.2           \\
FDU      & 7.8          & 17.2          & 37.5          & 47.3          & 65.1          & 96.7           & 6.3           & 13.7          & 29.1          & 36.3          & 49.0          & 71.1           & 11.8          & 27.2          & \textbf{41.3}          & 52.1          & 94.8          & 130.7           \\
AMHGCN   & 8.1          & 18.2          & 38.9          & 48.3          & 64.9          & 99.9           & 6.1           & 13.7          & 28.7          & 35.6          & 47.7          & 72.2           & 11.6          & 27.3          & 58.4          & 71.4          & 93.0          & 134.6           \\
SAGGB    & 7.4          & 17.1          & 37.8          & 47.9          & 65.7          & 101.9          & 6.2           & 14.1          & 29.8          & 37.3          & 51.1          & 75.1           & \textbf{10.9} & 26.8          & 59.8          & 73.9          & 96.9          & 137.5           \\
Ours     & \textbf{6.7} & \textbf{12.8} & \textbf{30.9} & \textbf{39.5} & \textbf{58.1} & \textbf{78.7}  
& 6.0  & \textbf{12.1} & \textbf{21.2} & \textbf{27.1} & \textbf{40.3} & \textbf{57.3}  
& 11.0 & \textbf{23.8} & 50.2 & \textbf{50.5} & \textbf{72.1} & \textbf{105.8}  \\ 
\hline
Time (ms) & 80           & 160           & 320           & 400           & 560           &                & 80            & 160           & 320           & 400           & 560           & 1000           & 80            & 160           & 320           & 400           & 560           & 1000            \\ 
\hline
         & \multicolumn{6}{c|}{Directions}                                                               & \multicolumn{6}{c|}{Sitting Down}                                                              & \multicolumn{6}{c}{Taking Photo}                                                                \\

HMR      & 23.3         & 25.0          & 47.2          & 61.5          & 80.9          & 116.9          & 9.6           & 18.6          & 41.1          & 57.7          & 101.7         & 148.3          & 7.9           & 19.0          & 31.5          & 57.3          & 83.5          & 108.5           \\
GA-MIN   & 6.8          & 15.3          & 42.1          & 50.2          & 68.1          & 100.0          & 14.5          & 25.5          & 56.3          & 70.3          & 90.8          & 135.2          & 8.3           & 16.6          & 38.2          & 49.0          & 59.7          & 115.3           \\
FDU      & 6.6          & 16.4          & 39.6          & 50.1          & 68.1          & 97.2           & 13.9          & 25.6          & 54.2          & 67.2          & 94.3          & 145.3          & 8.1           & 18.0          & 39.2          & 50.6          & 72.2          & 116.1           \\
AMHGCN   & 6.4          & 16.3          & 39.4          & 49.9          & 67.5          & 100.7          & 13.3          & 26.8          & 55.6          & 69.2          & 93.8          & 146.0          & 8.1           & 18.3          & 41.1          & 52.1          & 71.8          & 114.2           \\
SAGGB    & \textbf{6.2} & 16.0          & 39.0          & 50.0          & 70.6          & 101.8          & 12.8          & 26.3          & 55.9          & 70.3          & 96.9          & 150.2          & 7.8           & 17.9          & 41.3          & 52.9          & 77.3          & 118.6           \\
Ours     & 6.3          & \textbf{10.7} & \textbf{17.1} & \textbf{29.8} & \textbf{45.0} & \textbf{59.8}  & \textbf{8.9}  & \textbf{17.8} & \textbf{38.2} & \textbf{55.6} & \textbf{63.5} & \textbf{99.8}  & \textbf{7.8}  & \textbf{13.5} & \textbf{27.2} & \textbf{43.1} & \textbf{55.5} & \textbf{94.1}   \\ 
\hline
Time (ms) & 80           & 160           & 320           & 400           & 560           & 1000           & 80            & 160           & 320           & 400           & 560           & 1000           & 80            & 160           & 320           & 400           & 560           & 1000            \\ 
\hline
         & \multicolumn{6}{c|}{Sitting}                                                                  & \multicolumn{6}{c|}{Posing}                                                                    & \multicolumn{6}{c}{Greeting}                                                                    \\

HMR      & 12.6         & 25.6          & 44.7          & 60.7          & 76.4          & 118.4          & 13.6          & 23.5          & 62.5          & 114.1         & 126.3         & 143.6          & 12.9          & 31.9          & 55.6          & 82.5          & 104.3         & 123.2           \\
GA-MIN   & 8.1          & 18.5          & 41.9          & 53.2          & 70.2          & 109.1          & 7.8           & 19.3          & 43.4          & 56.0          & 83.2          & 150.2          & 12.8          & 26.3          & 61.8          & 75.8          & 95.3          & 121.5           \\
FDU      & 8.7          & 18.9          & 42.1          & 53.2          & 72.3          & 114.5          & \textbf{7.5}           & 19.3          & 47.1          & 62.0          & 93.3          & 149.5          & 13.0          & 30.7          & 63.1          & 78.2          & 109.4         & 141.8           \\
AMHGCN   & 8.5          & 18.7          & 42.3          & 53.7          & 73.8          & 115.8          & 9.7           & 24.7          & 60.6          & 77.8          & 108.1         & 169.2          & 13.4          & 32.1          & 70.3          & 85.8          & 109.1         & 146.3           \\
SAGGB    & \textbf{8.1} & 18.4          & 42.3          & 54.1          & 74.7          & 116.6          & 9.1  & 23.3          & 57.4          & 74.6          & 109.4         & 165.8          & 12.5          & 30.4          & 68.6          & 85.4          & 110.0         & 141.7           \\
Ours     & 8.2          & \textbf{16.1} & \textbf{37.2} & \textbf{48.9} & \textbf{78.5} & \textbf{108.4} & 9.6           & \textbf{18.8} & \textbf{45.2} & \textbf{50.0} & \textbf{65.1} & \textbf{108.2} & \textbf{9.2}  & \textbf{21.3} & \textbf{40.1} & \textbf{63.2} & \textbf{73.3} & \textbf{102.3}  \\ 
\hline
Time (ms) & 80           & 160           & 320           & 400           & 560           & 1000           & 80            & 160           & 320           & 400           & 560           & 1000           & 80            & 160           & 320           & 400           & 560           & 1000            \\ 
\hline
         & \multicolumn{6}{c|}{Waiting}                                                                  & \multicolumn{6}{c|}{Walking Dog}                                                               & \multicolumn{6}{c}{Average}                                                                     \\

HMR      & 12.8         & 24.5          & 45.2          & 85.1          & 87.5          & 121.9          & 30.1          & 41.4          & 78.4          & 100.1         & 134.7         & 157.4          & 14.5          & 25.0          & 49.5          & 72.6          & 93.2          & 123.5           \\
GA-MIN   & 7.5          & 17.2          & 41.1          & 52.3          & 70.7          & 102.2          & 18.9          & 38.5          & 70.9          & 84.0          & 104.1         & 138.2          & 10.1          & 21.5          & 47.2          & 58.4          & 76.6          & 115.8           \\
FDU      & 8.2          & 18.4          & 41.3          & 52.1          & 70.0          & 116.1          & 14.5          & 32.7          & 63.8          & 76.0          & 94.6          & 123.1          & 9.7           & 21.6          & 45.3          & 56.8          & 80.3          & 118.4           \\
AMHGCN   & 8.3          & 19.3          & 43.4          & 54.4          & 73.1          & 105.3          & 18.0          & 39.3          & 76.0          & 89.3          & 108.4         & 142.4          & 10.1          & 23.2          & 50.4          & 62.5          & 82.8          & 122.4           \\
SAGGB    & 7.6          & 17.9          & 41.1          & 52.3          & 73.3          & 104.1          & \textbf{16.0} & 36.0          & 72.0          & 85.2          & 103.8         & 137.3          & 9.5           & 22.2          & 49.5          & 62.2          & 84.5          & 122.8           \\
Ours     & \textbf{7.3} & \textbf{15.8} & \textbf{38.1} & \textbf{49.5} & \textbf{65.4} & \textbf{92.6}  & 16.1          & \textbf{32.1} & \textbf{63.0} & \textbf{70.4} & \textbf{98.4} & \textbf{109.2} & \textbf{8.8}  & \textbf{17.7} & \textbf{37.1} & \textbf{48.0} & \textbf{65.0} & \textbf{92.4}   \\
\hline
\end{tabular}}
\caption{Short-term and long-term prediction results on the H3.6M.}
\label{table1}
\end{table*}

\begin{table*}[h]
\centering
\resizebox{\textwidth}{!}{
\renewcommand\tabcolsep{6.0pt}
\begin{tabular}{c|cccccc|cccccc|cccccc} 
\hline
Time (ms)  & 80           & 160           & 320           & 400           & 560           & 1000          & 80           & 160           & 320           & 400           & 560           & 1000          & 80            & 160           & 320           & 400           & 560            & 1000            \\ 
\hline
           & \multicolumn{6}{c|}{Basketball}                                                              & \multicolumn{6}{c|}{Walking}                                                                 & \multicolumn{6}{c}{Directing Traffic}                                                            \\
Res-GRU    & 18.4         & 33.8          & 59.5          & 70.5          & 75.5          & 106.7         & 8.2          & 13.7          & 21.9          & 24.5          & 67.2          & 94.2          & 15.2          & 29.6          & 55.1          & 66.1          & 182.3          & 127.1           \\
FC-GCN     & 14.0         & 25.4          & 49.6          & 61.4          & 77.4          & 106.1         & 7.6          & 12.5          & 23.0          & 27.5          & 27.2          & 40.2          & 7.4           & 15.1          & 31.7          & 42.2          & 70.3           & 152.4           \\
GA-MIN     & 10.3         & 19.8          & 40.3          & 51.8          & —             & 88.8          & 5.2          & 8.9           & 16.2          & 18.2          & —             & 26.2          & 5.7           & 10.8          & 27.2          & 33.4          & —              & 137.8           \\
DPnet      & 10.7         & 17.8          & 38.4          & 49.5          & 58.1          & 98.4          & 5.8          & 9.0           & 17.2          & 21.4          & 24.9          & \textbf{34.1} & 5.9           & 11.8          & 26.6          & 33.5          & 66.6           & 143.3           \\
Ours       & 10.5         & \textbf{16.8} & \textbf{33.2} & 42.5          & \textbf{55.1} & \textbf{82.4} & \textbf{5.1} & \textbf{7.9}  & 15.2          & 18.2          & \textbf{23.1} & 40.2          & 4.3           & \textbf{9.2}  & \textbf{18.1} & \textbf{27.3} & \textbf{40.6}  & \textbf{99.1}   \\ 
\hline
Time (ms)  & 80           & 160           & 320           & 400           & 560           & 1000          & 80           & 160           & 320           & 400           & 560           & 1000          & 80            & 160           & 320           & 400           & 560            & 1000            \\ 
\hline
           & \multicolumn{6}{c|}{Soccer}                                                                  & \multicolumn{6}{c|}{Basketball Signal}                                                       & \multicolumn{6}{c}{Jumping}                                                                      \\
Res-GRU    & 20.3         & 39.5          & 71.3          & 84.0          & 79.7          & 129.6         & 12.7         & 23.8          & 40.3          & 46.7          & 65.5          & 77.5          & 36.0          & 68.7          & 125.0         & 145.5         & 132.3          & 195.5           \\
FC-GCN     & 12.1         & 21.8          & 41.9          & 52.9          & 82.6          & 117.5         & 3.5          & 6.1           & 11.7          & 15.2          & 25.3          & 53.9          & 22.4          & 44.0          & 87.5          & 106.3         & 131.4          & 164.6           \\
GA-MIN     & 9.8          & 18.3          & 39.0          & 49.4          & —             & 93.2          & 2.5          & 4.6           & 10.5          & 15.3          & —             & 58.3          & 14.2          & 28.2          & 71.8          & 91.1          & —              & 162.1           \\
DPnet      & 9.0          & 17.1          & 35.8          & 48.7          & 87.1          & 115.0         & 2.6          & \textbf{4.4}  & 10.0          & 13.4          & 30.1          & 61.2          & 12.4          & 28.3          & 70.2          & 89.2          & \textbf{100.1} & 166.1           \\
Ours       & \textbf{6.5} & \textbf{12.5} & \textbf{26.4} & \textbf{40.8} & \textbf{69.9} & \textbf{89.1} & \textbf{2.2} & 4.9           & \textbf{9.2}  & \textbf{10.1} & \textbf{18.9} & \textbf{45.2} & \textbf{10.6} & \textbf{20.4} & \textbf{51.8} & \textbf{80.2} & 109.1          & \textbf{140.2}  \\ 
\hline
Time (ms)  & 80           & 160           & 320           & 400           & 560           & 1000          & 80           & 160           & 320           & 400           & 560           & 1000          & 80            & 160           & 320           & 400           & 560            & 1000            \\ 
\hline
           & \multicolumn{6}{c|}{Wash Window}                                                             & \multicolumn{6}{c|}{Running}                                                                 & \multicolumn{6}{c}{Average}                                                                      \\
Res-GRU    & 8.4          & 15.8          & 29.3          & 24.5          & 85.3          & 102.7         & 25.8         & 48.9          & 88.2          & 100.8         & 124.5         & 158.1         & 18.1          & 34.2          & 61.3          & 70.3          & 101.5          & 123.9           \\
FC-GCN     & 5.9          & 11.9          & 19.4          & 23.1          & 53.0          & 79.3          & 26.0         & 36.6          & 38.8          & 39.5          & 26.1          & 58.2          & 12.4          & 21.7          & 38.0          & 46.0          & 61.7           & 96.5            \\
GA-MIN     & 4.5          & 9.9           & 27.8          & 35.2          & —             & 69.2          & 17.5         & 22.3          & 22.1          & 26.1          & —             & 40.1          & 8.7           & 15.4          & 31.9          & 40.1          & —              & 84.5            \\
DPnet      & 4.5          & 9.8           & 27.3          & 36.7          & 55.5          & 72.1          & 16.7         & 18.4          & 19.6          & 25.1          & \textbf{28.3} & 40.1          & 8.5           & 14.6          & 30.6          & 39.7          & 56.3           & 91.3            \\
Ours       & \textbf{4.2} & \textbf{8.2}  & \textbf{24.3} & \textbf{32.5} & \textbf{44.4} & \textbf{58.2} & \textbf{8.3} & \textbf{13.1} & \textbf{24.6} & \textbf{30.1} & 29.1          & \textbf{38.8} & \textbf{6.5}  & \textbf{11.6} & \textbf{25.4} & \textbf{35.2} & \textbf{48.8}  & \textbf{74.2}   \\
\hline
\end{tabular}}
\caption{Short-term and long-term prediction on the CMU MoCap.}
\label{table2}
\end{table*}

\section{Experiments}
In this section, we evaluate the proposed method on large benchmark datasets. We seek to answer the following research questions.
\textbf{Q1:} How is the proposed method comparing to state-of-the-art (SOTA) motion prediction approaches?
\textbf{Q2:} How do the visual results of the proposed method compare to SOTA motion prediction methods?
\textbf{Q3:} How much do different components of EPS contribute to its performance? 
\quad Then, we first present the experimental settings, followed by answering the above research questions.

\subsection{Datasets and Experimental Settings}
\quad Human 3.6 Million (\textbf{H3.6M}) dataset \cite{H36M} contains 3.6 million human images recorded by a Vicon motion capture system. 7 subjects perform 15 different classes of actions. Following the evaluation protocol of previous work \cite{srnn}, duplicate points in poses are removed and downsampled to 25 FPS. S5 is utilized as the test set.
\quad CMU Motion Capture (\textbf{CMU Mocap}) dataset is released by researchers from Carnegie Mellon University. 12 infrared cameras are utilized to capture human poses. Following previous works \cite{LTD}, we adopt the same training/test splits.
\quad 3D Poses in the Wild (\textbf{3DPW}) dataset \cite{3Dpose} is proposed primarily for wild scenes that are recorded by a handheld smartphone camera or IMU. It contains 60 video sequences with more than 51,000 indoor or outdoor poses. 

We build our model on the PyTorch with a NVIDIA 3090Ti GPU. The Adam Optimizer is utilized with a learning rate of 0.001. The model is trained for 15 epochs with a batch size of 16. The loss weight $\alpha_1=\alpha_2=1$, $\beta_1=0.9$, and $\beta_2=0.1$.
We evaluate our proposed approach by measuring the mean per joint position error (MPJPE) after the alignment of the root joint. Experimental results at 80 $ms$, 160 $ms$, 320 $ms$, 400 $ms$, and 1,000 $ms$ in the future are shown for comparisons.


\subsection{Comparisons with Existing Approaches (RQ1)}
\textbf{Results on H3.6M.}\quad
We compare our method with SOTA approaches: AMHGCN \cite{AMH} and HMR \cite{HMR} (RNN-based); FDU \cite{FDU} (graph-based); GA-MIN \cite{GAmin} (DCT-Transformer hybrid); and SAGGB \cite{SAGGB} (temporal-graph convolutional). These methods are retrained using the authors' official implementations to ensure fair evaluation. For a detailed introduction of the relevant works, please refer to the supplementary documents.

As shown in Table~\ref{table1}, our method demonstrates substantial improvements across both prediction horizons. 
\quad \textbf{\textit{For short-term prediction}} (0-400 $ms$), APS achieves an average improvement of 7.3-46.2\% over baselines, with particularly significant gains of 12.8-39.5\% against SAGGB. The effectiveness in capturing spatio-temporal patterns is especially evident in complex motions, where APS reduce errors by 29.8-73.8\% compared to SAGGB.
\quad \textit{\textbf{For long-term prediction}} (400-1,000$ms$), our method maintains superior performance with 17.1-59.8\% lower errors on average. Notably, we outperform SAGGB by 19.9-44.7\% across all time horizons, with particular results (31.6-59.8\%) for challenging sequences (\textit{e.g.,} Directions). The consistent accuracy across different prediction windows confirms the robustness of APS in modeling long-term dependencies.
Quantitative results demonstrate our method superior performance in both prediction horizons, with strong results for non-periodic motions (e.g., 59.8 $mm$ vs 101.8 $mm$ error for 'Directions' at 1,000 $ms$) compared to periodic ones (e.g., 109.2 $mm$ vs 137.3 $mm$ for 'Walking Dog'). This performance gap suggests our quotient space modeling is especially effective for complex motions.


\textbf{Results on CMU Mocap.}\quad
To verify the effectiveness and generalization of APS, experiments are carried out on the CMU Mocap, which has a more complex action performance than the H3.6M, and the results are detailed in Table~\ref{table2}. Existing advanced methods are compared with APS, including Res-Gru \cite{Res-gru}, FC-GCN \cite{LDR}, GA-MIN \cite{GAmin}, and DPnet \cite{DPnet}.
\quad It can be observed that our method performs optimally for \textbf{\textit{the short-term}}, and 11.2\% (400 $ms$) in comparison to the DPnet. Compared to the DPnet, APS improves 23.5\% (80 $ms$) and 11.2\% (400 $ms$) on average prediction accuracy. 
\quad \textit{\textbf{In terms of long-term prediction}}, our method shows significant accuracy improvement of 18.7\%-40.0\%. The experimental results demonstrate that the spatio-temporal perception enhancement model can effectively capture the spatio-temporal dependence under longer pose sequences (400-1,000 $ms$). In addition, our method shows optimal prediction results on two major datasets, which further proves the effectiveness and robustness.

\begin{table}[t]
\centering
\resizebox{.46\textwidth}{!}{
        \renewcommand\tabcolsep{10.0pt}
	\begin{tabular}{cccccccccccc}
	\hline
	Time (ms) & 200 & 400 & 600 & 800 & 1000  \\ \hline
        Res-GRU      & 37.3 & 67.8 & 94.5 & 109.7 & 123.6 \\ 
	DMGNN      & 37.3 & 67.8 & 94.5 & 109.7 & 123.6 \\ 
        MSR-GCN     & 37.8 & 71.3 & 93.9 & 110.8 & 121.5 \\
	FDU     & 26.1 & 54.2 & 72.3 & 87.2 & 94.5\\ 
	 Ours       & \textbf{19.0} & \textbf{47.5} & \textbf{67.8} & \textbf{79.3} & \textbf{89.3}\\ \hline
	\end{tabular}}
\caption{Average prediction errors on 3DPW.}
\label{tab3}
\end{table} 

\textbf{Results on 3DPW.}\quad
The 3DPW dataset poses significant challenges due to its capture of human motions in uncontrolled environments using handheld smartphones. As shown in Table~\ref{tab3}, our method establishes new SOTA performance across all prediction horizons. Specifically, at 200 $ms$, our method achieves an error of 19.0 $mm$, a 27.2\% improvement over FDU (26.1 $mm$). This advantage becomes more pronounced for longer-term predictions, reducing the error from 94.5 $mm$ to 89.3 $mm$ (5.5\%) at 1,000 $ms$. Our method maintains consistent superiority, demonstrating 49.1\% lower error than Res-GRU (123.6 $mm$) and 26.5\% lower than MSR-GCN (121.5 $mm$) at 1,000 $ms$. With an average improvement of 10. 2\% over FDU, these results validate the robustness in handling real-world motion variability in three datasets.

\begin{figure*}
	\centering
	\includegraphics[width=.9\textwidth]{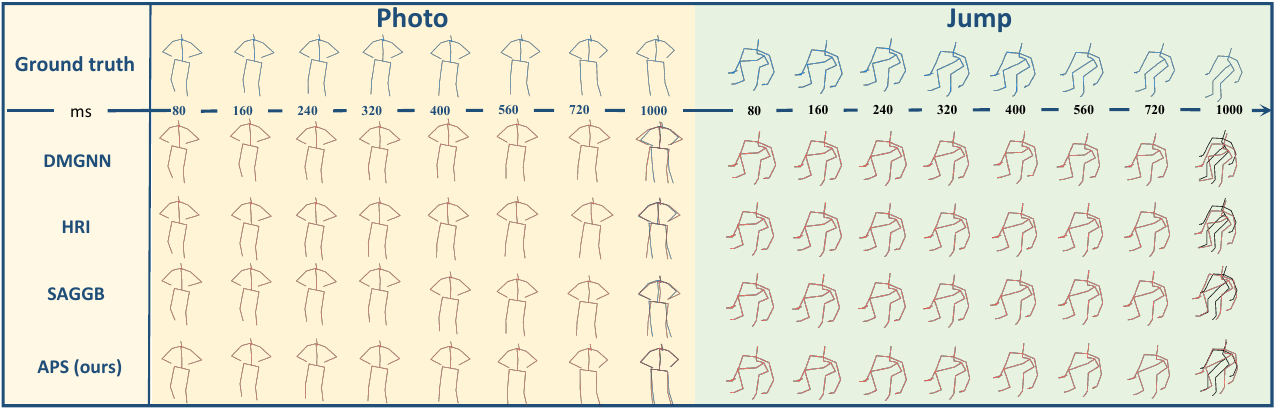}  
	\caption{ Visual results of our model on H3.6M and CMU datasets.}
	\label{fig:vision}
\end{figure*}

\textbf{Further analysis.}\quad 
Our perceptual strategy achieves success through multiple synergistic advantages.  By employing physically interpretable operators, it decouples implicit motion features from raw pose data, reducing the feature extraction burden on the network. The quotient space reduces high-dimensional pose space to low-dimensional equivalence classes, while its projection angle characteristics naturally align with anatomical planes to enable effective motion pattern separation. The differential modeling approach suppresses static displacement artifacts while maintaining dynamic sensitivity for a precise motion response. Through active joint information modification, the method enforces meaningful spatiotemporal relationship learning.

\subsection{Visual Experiments (RQ2)}
As illustrated in Figure~\ref{fig:vision}, we evaluate our method on the H3.6M dataset and compare it with three SOTA approaches: DMGNN, HRI, and SAGGB. We select the "Photo" action for analysis, as it engages all four limbs, providing a comprehensive test of spatiotemporal feature capture. Our method produces predictions that align more closely with ground-truth poses than competing approaches. Notably, this action involves subtle motions, which often cause models to prematurely converge to static states. Our approach maintains accurate motion dynamics over extended horizons, demonstrating superior spatiotemporal dependency modeling. This robustness comes from our method's ability to actively learn motion relations through controlled joint perturbation.
Figure~\ref{fig:vision} further presents results on the CMU Mocap dataset for the "Jump" action, characterized by large spatial displacements. For clarity, we overlay each predicted pose with a semi-transparent ground-truth reference. Both short- and long-term predictions from our method exhibit significantly tighter alignment with real sequences compared to alternatives. Crucially, our model accurately tracks subtle upper-body motions despite aggressive lower-limb movements, underscoring its resilience to amplitude variations. 
This empirically validates the effectiveness in decoupling and reconstructing motion patterns.



\begin{table}[t]
\centering
\resizebox{.48\textwidth}{!}{
\renewcommand\tabcolsep{14.0pt}
\begin{tabular}{ccc|ccccc}
\hline
 D & E & L & 80   & 160  & 320  & 400  & 1000  \\ \hline
  & \checkmark & \checkmark & 10.5 & 25.6 & 45.4 & 58.3 & 110.5 \\
\checkmark &   & \checkmark & 8.5  & 20.1 & 41.0 & 55.3 & 98.3  \\
\checkmark & \checkmark &   & 8.6  & 19.7 & 39.3 & 50.7 & 93.5  \\
\checkmark & \checkmark & \checkmark & 8.8  & 17.7 & 37.1 & 48.0 & 92.4  \\ \hline
\end{tabular}}
\caption{Ablation study results.}
\label{table4}
\end{table}

\subsection{Ablation Experiments (RQ3)}

To rigorously evaluate our components of APS, we conduct an ablation study on H3.6M (Table \ref{table4}), analyzing three modules: the \textbf{\textit{D}}ata Perceptual Module (D), which processes motion in quotient space; the Spatio-Temporal \textbf{\textit{E}}nhancement Component (E), refining joint correlations; and the Spatio-Temporal \textbf{\textit{L}}earning Component (L), featuring our novel attention mechanism. 
Key findings emerge:
Replacing quotient space with conventional 3D poses (removing D) degrades performance significantly ($e.g.,$ 110.5 $mm$ vs. 92.4 $mm$ at 1,000 $ms$), confirming its efficacy in preserving dynamics while reducing dimensionality.
Removing D (retaining E+L) causes notable drops, especially in medium-term predictions ($e.g.,$ 20.1 $mm$ vs. 17.7 $mm$ at 160 $ms$), validating our joint perturbation-recovery strategy for capturing spatio-temporal relationships.
Substituting L with a standard Transformer (keeping D+E) leads to consistent degradation, underscoring the superiority of our decoupled spatio-temporal attention over joint modeling approaches.
The whole model (D+E+L) achieves optimal performance ($e.g.,$ 92.4 $mm$ at 1,000 $ms$), demonstrating the complementary roles of the modules. These results systematically validate our design choices, revealing how each component addresses distinct challenges in motion prediction.

\begin{table}[t]
	
\centering
	\resizebox{.5\textwidth}{!}{
		\renewcommand\tabcolsep{6.0pt}
		\begin{tabular}{ccccccccccccccccccccc}
			\hline
			& \multicolumn{5}{c}{Current methods}    & \multicolumn{5}{c}{Current methods + APS}  \\ \hline
			Time (ms) & 80 & 160 & 320 & 400 & 1000  & 80 & 160 & 320 & 400 & 1000   \\ \hline
			BiGAN   &13.6& 26.1& 51.4& 63.1& 84.1& 11.5& 25.5& 46.2& 58.8& 83.5 \\ 
			LTD  &7.7& 15.8& 30.5& 37.6&74.1 &7.1& 14.9& 31.0& 36.0& 72.0 \\ 
                FDU   &6.3& 13.7& 29.1& 36.3& 59.3&5.7& 12.6& 26.2& 35.7& 56.1 \\ 
               SAGGB  &6.2& 14.1& 29.8& 37.3&57.3& 5.3& 12.0& 26.2& 32.3& 50.5\\
			
			\hline
			
	\end{tabular}}
\caption{Studies on model-agnostic learning.}
 \label{tab:srnn}
\end{table}

\subsection{Studies on Model-agnostic Learning}
Our proposed APS is a model-agnostic framework, which can be easily integrated with existing methods. To demonstrate its effectiveness, we conduct experiments by utilizing various well-established methods such as LTD \cite{LTD}, DMGNN \cite{DMGNN}, MSR-GCN \cite{MSGNN}, and FDU \cite{FDU}. These methods represent different types of networks including RNNs, GCNs, and GANs, thereby covering a wide range of network classes. The results, as shown in Table \ref{tab:srnn}, clearly indicate that APS significantly improves the prediction accuracy of these existing methods. 
Notably, APS mitigates the performance degradation typically observed in long-term predictions (\textit{$e.g.,$} 1,000 $ms$). For instance, SAGGB+APS achieves a 12\% lower error (50.5 $mm$ vs. 57.3 $mm$ ) at 1,000$ms$, while FDU+APS reduces errors by 5.4\% (56.1 $mm$  vs. 59.3 $mm$ ). 
Crucially, the framework delays performance saturation by enhancing temporal coherence and error correction, which is especially vital for complex motion dynamics. These results validate APS as a universally applicable tool that not only boosts accuracy but also extends the effective prediction range of diverse architectures, addressing key limitations in HMP.

\section{Conclusion}
We introduce an Active Perceptual Strategy (APS) for 3D human motion prediction, addressing the issue through two key innovations:
\quad i) APS actively separates motion dynamics from coordinate redundancy via quotient space projection (Grassmann manifold and tangent vectors), compelling the network to focus on fundamental kinematic constraints and semantic motion.
\quad ii) We design an auxiliary training signal through adversarial perturbation, forcing the network to actively recover corrupted spatiotemporal relationships, thereby overcoming the limitations of passive data fitting.
Our framework shifts the conventional "black-box end-to-end fitting" paradigm to an active approach combining "disentangled representation" with "adversarial refinement". Extensive results on human motion datasets demonstrate the competency of our approach.

\section{Acknowledgments}
This work is supported by the National Key Research and Development Program, Unified Integrated Development technology of scientific computing language and engineering physics modeling language (Grant No.2024YFB3310200), and the Key R\&D plan of Jilin Province (Grant No. 20250201076GX).

\bibliography{aaai2026}

\end{document}